\pdfoutput=1
\documentclass[11pt]{article}

\usepackage{acl}

\usepackage{times}
\usepackage{latexsym}

\usepackage[T1]{fontenc}
\usepackage[utf8]{inputenc}

\usepackage{graphicx}

\usepackage{microtype}

\usepackage{tabularx}
\usepackage{footnote}\makesavenoteenv{tabular}
\usepackage{float}
\usepackage{rotating,booktabs,multirow}
\usepackage{amsmath,amsfonts,amssymb}
\usepackage{adjustbox}
\usepackage{makecell} 
\usepackage[shortlabels]{enumitem}

\usepackage{CJKutf8}
\usepackage{pdfpages}
\usepackage{xcolor}
\usepackage[normalem]{ulem}               % to striketrhourhg text
\newcommand\redout{\bgroup\markoverwith
{\textcolor{red}{\rule[0.5ex]{2pt}{0.8pt}}}\ULon}

\usepackage{cancel}

\usepackage{array}
\usepackage{color, colortbl}

\usepackage{listings}

\newcommand\ignore[1]{}

\definecolor{burntorange}{rgb}{0.8, 0.33, 0.0}

\title{Investigating Math Word Problems using Pretrained Multilingual Language Models}

\author{First Author \\
  Affiliation / Address line 1 \\
  Affiliation / Address line 2 \\
  Affiliation / Address line 3 \\
  \texttt{email@domain} \\\And
  Second Author \\
  Affiliation / Address line 1 \\
  Affiliation / Address line 2 \\
  Affiliation / Address line 3 \\
  \texttt{email@domain} \\}
  
\author{Minghuan Tan \and Lei Wang \and Lingxiao Jiang \and Jing Jiang \\
        School of Computing and Information Systems \\
        Singapore Management University \\
        \{mhtan.2017,lei.wang.2019\}@phdcs.smu.edu.sg,\{lxjiang,jingjiang\}@smu.edu.sg}

\begin{document}
\maketitle
\begin{abstract}
In this paper, we revisit math word problems~(MWPs) from the {\em cross-lingual} and {\em multilingual} perspective.
We construct our MWP solvers over pretrained multilingual language models using the sequence-to-sequence model with copy mechanism.
We compare how the MWP solvers perform in cross-lingual and multilingual scenarios.
To facilitate the comparison of cross-lingual performance, we first adapt the large-scale English dataset MathQA as a counterpart of the Chinese dataset Math23K.
Then we extend several English datasets to bilingual datasets through machine translation plus human annotation.
Our experiments show that the MWP solvers may not be transferred to a different language even if the target expressions share the same numerical constants and operator set.
However, it can be better generalized if problem types exist on both source language and target language.
% \lx{may be more interesting to give some concrete results before saying the limit.}
\end{abstract}
\begin{CJK*}{UTF8}{gbsn}

\section{Introduction}

\begin{figure}[!t]
\centering
\includegraphics[width=\linewidth]{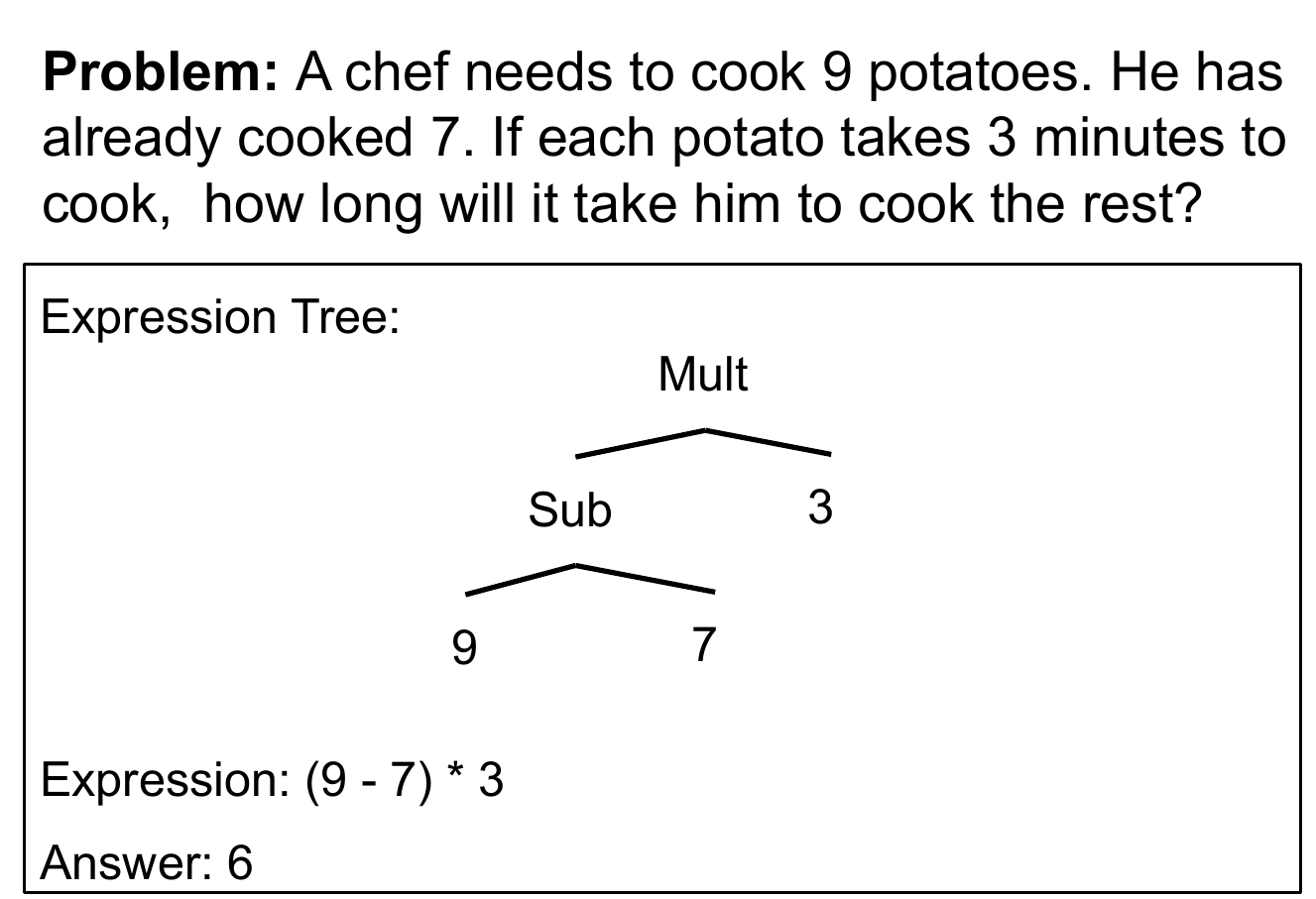}
\caption{Example of an MWP and its expression tree.} 
\vspace{-5mm}
\label{fig:example}
\end{figure}

How to use machine learning and NLP techniques to solve Math Word Problems~(MWPs) has attracted much attention in recent years~\cite{hosseini-etal-2014-learning,kushman-etal-2014-learning,roy-etal-2015-reasoning,ling-etal-2017-program,wang-etal-2017-deep-neural,wang-etal-2018-translating,amini-etal-2019-mathqa}.
% \jjcomment{cite papers}
Given a math problem expressed in human language, a MWP solver typically first converts the input sequence of words to an \emph{expression tree} consisting of math operators and numerical values, and then invokes an executor (such as the \emph{eval} function in Python) to execute the expression tree to obtain the final numerical answer.
Figure~\ref{fig:example} shows an example math word problem, the correct expression tree, and the final answer.
% \jjcomment{include a figure to illustrate the problem, for readers who're not familiar with the MWP solver task.}
% This task therefore can be seen as a special case of \emph{code synthesis}~\cite{desai2016program,lin-etal-2018-nl2bash,Shin2019,wang-etal-2020-rat} in a restricted domain, where the code to be synthesized is the expression tree.
% \jjcomment{This sentence is to link this paper to the core topics covered by the workshop. Maybe you can ask Prof. Lingxiao what to cite for code synthesis. I found that code synthesis seems to be the topic explicitly mentioned by the workshop that's closest to your work.}
% \lx{May cite sample papers for different kinds of code synthesis from UCB https://sunblaze-ucb.github.io/program-synthesis/index.html or MSR https://www.microsoft.com/en-us/research/project/program/
% 	E.g., PATOIS~\cite{Shin2019}, RAT-SQL~\cite{Wang2020}, a couple of NL-to-DSL synthesizers~\cite{desai2016program}, NL2Bash~\cite{Lin2018}, etc. But note that Big Code != Big Vocabulary \cite{Karampatsis2020}, out-of-vocabulary words may affect the models a lot.
% }

Despite the relatively simple syntax of these expression trees, building MWP solvers is not a trivial task, and researchers have proposed various methods to tackle the different challenges of this problem such as statistical methods~\cite{kushman-etal-2014-learning,roy-etal-2015-reasoning}, parsing-based methods~\cite{shi-etal-2015-automatically} and generation-based methods~\cite{wang-etal-2018-translating,xie-etal-ijcai-2019-goal}.
% \jjcomment{I intentionally kept this part short because we don't want to go into great details of existing work on general MWP solving.}
However, an aspect that has been largely overlooked is cross-lingual and multilingual MWP solving, i.e., whether a MWP solver trained on one human language can still work on another human language, or whether a MWP solver trained on multiple human languages together is more effective than a solver trained on only one language.
% \jjcomment{I think it's important to describe what we mean by cross-lingual and multilingual MWP solving at this point.}
We believe this is an interesting aspect to study for the following reasons.
First, in cognitive science, people have long studied the relationship between humans' numerical processing abilities and language abilities, and found that on the one hand, the two are largely independent~\cite{xu2000large}, but on the other hand, ``acquiring and mastering symbolic representations of exact quantities critically depends on language and instruction"~\cite{van2015relation}.
% \jjcomment{The previous sentence is to cite studies and their main findings from cognitive science (or other related disciplines) that tries to understand the interplay between humans' language abilities and numeracy abilities.}
It is therefore also intriguing to study whether machines separately acquire arithmetic and language abilities.
% \jjcomment{This previous sentence probably needs to be revised to precisely describe what we try to understand.}
Second, with pre-trained large-scale multilingual language models such as mBERT~\cite{devlin-etal-2019-bert} and XLM-R~\cite{conneau-etal-2020-unsupervised}, which presumably project different human languages into a common embedding space, we have seen some success in cross-lingual NLP tasks such as XNLI~\cite{conneau-etal-2018-xnli} and MLQA~\cite{lewis-etal-2020-mlqa} in both zero-shot and few-shot settings~\cite{wu-dredze-2019-beto,conneau-etal-2020-unsupervised}.
% \jjcomment{By citing success in other NLP tasks that use pre-trained multilingual language models, we hope that it's easy to convince the readers that it is natural to also consider cross-lingual and multilingual MWP solving using these pre-trained models.}
It is therefore reasonable to expect that for MWP solving, there is the possibility of transferring machine's capability of MWP solving from one language to another by leveraging these pre-trained multilingual language models.

In this paper, we conduct an empirical study to understand to what extent MWP solvers can work in cross-lingual and multilingual settings.
% \jjcomment{This previous sentence positions our work. It is important to get this sentence right, i.e., to state it clearly and to state it based on what we believe this work is about. If you think it should be phrased differently, you should modify it. I wrote this sentence based on our understanding of what this work is about. (E.g., I didn't position the work as something like ``we study how to improve cross-lingual MWP solving by proposing a novel solution that ...''.)}
Specifically, we ask the following questions:
% \jjcomment{Here you can list the specific research questions your experiments try to answer.}
(1) Cross-lingual setting: Given a model trained with monolingual dataset, can the model solve MWPs over another language?
(2) Multilingual setting: Can combining datasets of different languages further boost the performance for each language?
(3) Can we identify some critical factors that may affect the results in (1) and (2)?

In order to empirically answer the questions above, we need multilingual MWP datasets, which are limited currently.
We first use large scale datasets like Math23K~\cite{wang2017deep} and MathQA~\cite{amini-etal-2019-mathqa} as monolingual MWPs resource and further adapt MathQA to have the same operator set and expression style with Math23K.
To better evaluate the models with parallel corpus, we extend some existing MWP datasets by translating them from English into Chinese. 
% The evaluation datasets are split according to the type and number of operators  used in the target expressions. 
% \jjcomment{Briefly describe what you've done with dataset construction.}
We then conduct three sets of experiments on the constructed datasets.
We find that: (1) a cross-lingual MWP solver finetuned on one language cannot work on a second language, even if they are sharing the same decoding vocabulary, (2) a multilingual MWP solver may not boost performance for all the training languages but can improve those problems of similar types if one training language is close to the evaluation language, (3) combining (1) (2), we think for multilingual MWP solvers, despite language similarity, the performance relies heavily on domain similarity~(problem types).
% xxx \jjcomment{summarize the major findings}.

Our work makes the following contributions:
(1) To the best of our knowledge, we are the first to study cross-lingual and multilingual MWP solving, and we empirically demonstrate that cross-lingual MWP solving is still difficult, but multilingual MWP solving is to some extent effective. 
(2) We discover that multilingual MWP solving is mostly effective for questions with similar problem types.
(3) Our constructed datasets can help other researchers to further study cross-lingual and multilingual MWP solving.

\section{Related work}

% \subsection{Math Word Problems Solvers}
Solving Math Word Problems~(MWPs) has been attracting researchers since the emergence of artificial intelligence.
STUDENT\cite{bobrow-daniel-1964-natural} is a rule-based math word problem solver which contains a pipeline that consists of heuristics for pattern transformation.
Many researchers start with the fundamental problem types like addition and subtraction~\cite{hosseini-etal-2014-learning} or those that have only one single operator~\cite{roy-etal-2015-reasoning}.
\citet{roy-roth-2015-solving} look at problems that require multi-steps using two or more operators.
The question types of MWPs are also expanding.
Rather than focusing on problems that need only one variable, \newcite{kushman-etal-2014-learning} propose a dataset ALG514 which includes problems with a system of equations.
With the development of deep learning, there has been a demand for large-scale datasets with more variations.
Dolphin18K~\cite{huang-etal-2016-well} is a large-scale dataset that is more than 9 times of the size of previous ones, and contains many more problem types.
Math23K~\cite{wang-etal-2017-deep-neural} contains math word problems for elementary school students in Chinese language and is crawled from multiple online education websites.
MathQA~\cite{amini-etal-2019-mathqa} is a new large-scale, diverse dataset of 37k multiple-choice math word problems in English and each problem is annotated with an executable formula using a new operation-based representation language.
HMWP~\cite{qin-etal-2020-semantically} contains three types of MWPs: arithmetic word problems, equations set problems, and non-linear equation problems. 

Various approaches have been proposed to solve MWPs.
% In this paper, we are more interested in how each method may use the input data, especially the numerical values.
Template-based approaches~\cite{kushman-etal-2014-learning,zhou-etal-2015-learn,upadhyay-etal-2016-learning,huang-etal-2017-learning} are widely adopted as numbers appeared in the expressions are usually sparse in the representation space and the expressions may fall into the same category.
More recently, the community is also paying more attention to train a math solver by fine-tuning pretrained language models.
For example, EPT~\cite{kim-etal-2020-point} adopts ALBERT~\cite{Lan2020ALBERT} as the encoder for its sequence-to-sequence module.

%From our point of view, 
The monolingual performance gains achieved recently have not been evaluated from cross-lingual and cross-domain perspectives.
Therefore, we decide to revisit MWPs using current SOTA pretrained multilingual language models to construct a competitive math solver and conduct experiments over various bilingual evaluations.

\section{Preliminaries}

\subsection{The MWP Solver Task}

We first formally define the task of building MWP solvers.
Given a math word problem with $n$ words $W=(w_1, w_2, \ldots, w_n)$, and $k$ numerical values $N=(n_0, n_1, \dots, n_k)$, the model needs to generate a flattened tree representation using operators from permitted operator set $\mathcal{O}$ and numerical values from constants $\mathcal{C}$ and $N$.
The generated tree should be able to be evaluated via some compiler and executor to return a numerical value.

\subsection{Solution Framework}

A MWP solver needs to generate executable code for a target programming language to be evaluated by an executor compiled for the programming language.

Our MWP solver is built upon a sequence-to-sequence model with copy mechanism~\cite{gu-etal-2016-incorporating}.
% \cite{kim-etal-2020-point}
Specifically, we use a pretrained multilingual model as the encoder to get contextualized representations of math word problems.
Due to the word piece tokenizer, the encoded context is not well-aligned to original input words.
We choose to map these word pieces back to input words through mean pooling.
Then we pass the mean pooled word representations to a bidirectional LSTM.
Finally, we use a LSTM decoder with copy mechanism, which takes in the last decoded word vector and intermediate reading states, to predict the next token one by one. 
When the decoding finishes, we are expecting to get a linear tree representation.
We attach the full model details in Section~\ref{sec:method}.

Given the decoded tree representation, we first convert the generated linear tree representation into a piece of python expression with basic operators (+,-,*,/,**), then use the built-in function \textit{eval} in Python to execute the generated code.

\subsection{Existing Datasets}
\label{subsec:datasets}

We use two large-scale datasets for this cross-lingual research.
One is Math23K~\cite{wang-etal-2017-deep-neural} in Chinese and the other is MathQA~\cite{amini-etal-2019-mathqa} in English.
Although the two datasets are similar in size and question types, there are still differences in terms with permitted operators and annotations.

\paragraph{Math23K}  The dataset Math23K~\cite{wang-etal-2017-deep-neural} contains math word problems for elementary school students in \emph{Chinese}~(zh) and is crawled from multiple online education websites.
The dataset focuses on arithmetic problems with a single-variable and contains 23,161 problems labeled with structured equations and answers. 

\paragraph{MathQA} The dataset is a new large-scale, diverse dataset of 37k multiple-choice math word problems in \emph{English}~(en). 
Each question is annotated with an executable formula using a new operation-based representation language~\cite{amini-etal-2019-mathqa}.
It covers multiple math domain categories.
To make MathQA a comparable counterpart with Math23K, we choose to filter those solvable problems with shared permitted operators from MathQA to create an adapted MathQA dataset.

\paragraph{Other datasets focusing on specific problem types} These datasets are smaller in size but more focused on specific problem types.
We follow the dataset naming conventions from MAWPS~\cite{koncel-kedziorski-etal-2016-mawps}. 

\begin{table}[t!]
    \centering
    \begin{adjustbox}{max width=\linewidth}
    \begin{tabular}{llr}\toprule
    Dataset &Problem Types &Size \\\midrule
    AddSub~\cite{hosseini-etal-2014-learning} &Add &395 \\
    &Sub & \\
    SingleOp~\cite{roy-etal-2015-reasoning} &Add &562 \\
    &Sub & \\
    &Mult & \\
    &Div & \\
    MultiArith~\cite{roy-roth-2015-solving} &(Add, Sub) &600 \\
    &(Sub, Add) & \\
    &(Add, Mult) & \\
    &(Add, Div) & \\
    &(Sub, Mult) & \\
    &(Sub, Div) & \\
    \bottomrule
    \end{tabular}
    % \end{small}
    \end{adjustbox}
    \caption{Datasets which are focusing on specific problem types.}
    \label{tab:evaluation}
\end{table}

Specifically, AddSub~\cite{hosseini-etal-2014-learning} covers arithmetic word problems on addition and subtraction for third, fourth, and fifth graders. 
Its problem types include combinations of additions, subtractions, one unknown equation, and U.S. money word problems. 
SingleOp~\cite{roy-etal-2015-reasoning} is a dataset with elementary math word problems of single operation.
MultiArith~\cite{roy-roth-2015-solving} includes problems with multiple steps which we listed all the seven types in Table~\ref{tab:evaluation}. 
% ~\leicomment{Is it necessary to state these datasets are English datasets not Chinese.}
These datasets are all in \emph{English}~(en).
We will illustrate how we extend them into bilingual datasets in Section~\ref{subsec:bilingual}.

\section{Cross-lingual and Multilingual MWP Solvers}

In this work, as we are focusing on the cross-lingual and multilingual properties of MWPs, we need to train separate MWP solvers using different datasets.
Our cross-lingual MWP solver will be trained using one language but evaluated using another.
Our multilingual MWP solver can be trained on all languages available and evaluated separately.
To suffice these goals, it would be better if the problems in different languages have comparable properties.
Since we are using pretrained multilingual language models as the sequence embedder of the encoder, all the languages can be projected into a shared representation.
However, the candidate datasets also need to share a common operator set and numerical constants to make the decoding process consistent.
But some of the categories from MathQA do not exist on Math23K or one of the operators is not in our permitted set.
Therefore, we need to adapt MathQA as a counterpart of Math23K sharing the same decoding vocabulary, including operators and constants.

\begin{figure}[ht]
\centering
\includegraphics[width=\linewidth]{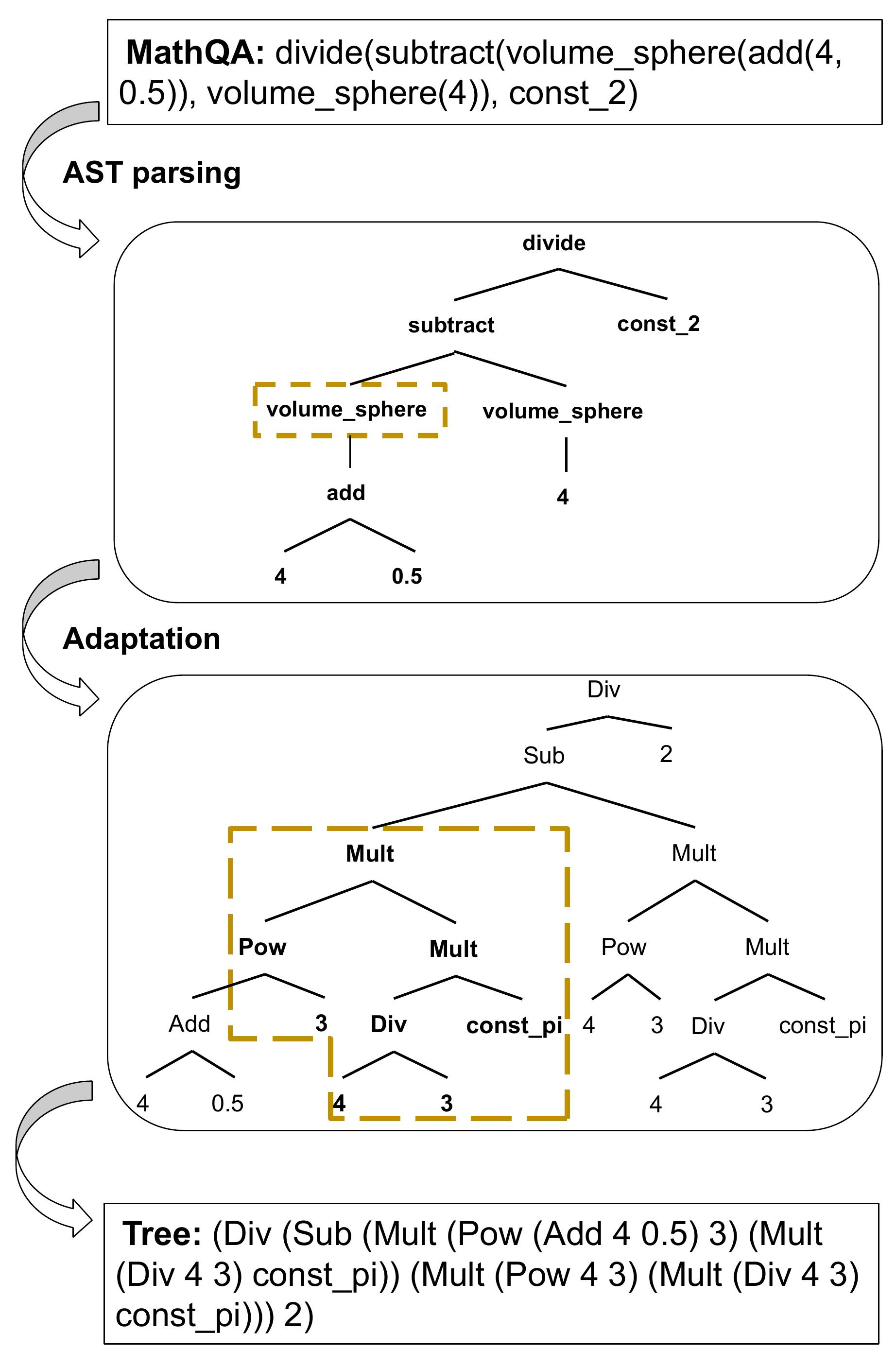}
\caption{Adaptation of MathQA to Math23K. The part highlighted with dashed lines shows the adaptation of the function \emph{volume\_sphere}.} 
\label{fig:adaptation}
\end{figure}

\subsection{Adaptation of MathQA}
% ~\leicomment{A bit confused about the title. what do the cross-lingual MWPs mean?}
We adapt MathQA by doing the following:

\begin{enumerate}[1),leftmargin=1em,nosep]
    \item We notice that the annotated formulas in MathQA are function calls of predefined functions which can be converted into a tree using an abstract syntax tree~(AST) parser.
    \item To be consistent with Math23K, which covers only basic arithmetic operators like addition~(Add), subtraction~(Sub), division~(Div), multiplication~(Mult) and exponentiation~(Pow), we keep only functions in MathQA that can be expressed in such operators.
    For example, $\mathit{volume\_sphere}(r)$ from MathQA equalizes to $\frac{4}{3}\pi r^3$ and is adapted using the method shown in Figure~\ref{fig:adaptation}.
    Formulas containing operators not used in Math23K, like \textit{sine} and \textit{permutation}, are not considered in this work.
    A full list of adapted operators can be found in Table~\ref{tab:operators} of Appendix~\ref{sec:method}.
    
    \item Upon constructing the trees using permitted operators, we evaluate each sample to verify its correctness against its ground-truth answer.
    % \lx{since the 'method' section is removed, it became unclear how is the 'evaluate' done; may add a sentence to briefly explain how the built-in function \textit{eval} in Python is used to calculate the answer.}
    Those cases that fail to get the correct answer are not considered in this work.
\end{enumerate}

\begin{table}[!htp]
    \centering
    % \begin{adjustbox}{max width=\textwidth}
    % \begin{small}
    \scalebox{0.86}{
\begin{tabular}{lrrrrrrr}\toprule
& &\multicolumn{2}{c}{Math23K} & &\multicolumn{2}{c}{MathQA} \\\cmidrule{3-4}\cmidrule{6-7}
& &w/o Pow &w/ Pow & &w/o Pow &w/ Pow \\\midrule
Train & &21,107 &21,161 & &15,302 &16,645 \\
Dev & &995 &1,000 & &2,263 &2,479 \\
Test & &999 &1,000 & &1,532& 1,653 \\
\bottomrule
    \end{tabular}}
    % \end{small}
    % \end{adjustbox}
    \caption{Statistics of different splits for Math23K and the adapted MathQA.
    % {\color{blue}let this two-column table be one-column table}
    }
    \vspace{-5mm}
    \label{tab:train}
\end{table}

% \subsection{Comparison of the Adapted MathQA and Math23K}
After the adaptation, we get the adapted MathQA dataset of solvable problems with comparable sizes and question types to Math23K.
For Math23K, we further sample a development set of size 1000 from its training set.
Considering the operator Pow has only several training and evaluating instances on Math23K, we separate them with others to make a fairer adaptation of MathQA to Math23K. 
We show the statistics of both Math23K and the adapted MathQA in Table~\ref{tab:train}.
In this work, all the experiments will be conducted on the dataset marked with w/o Pow.

\begin{table*}[ht]
    \centering
    \scalebox{0.86}{
    \begin{tabular}{lp{4cm}p{4cm}p{6cm}}\toprule
    Dataset &\multicolumn{1}{c}{AddSub} & \multicolumn{1}{c}{SingleOp }& \multicolumn{1}{c}{MultiArith} \\\midrule
    Problem Types & \multicolumn{1}{c}{addition, subtraction} &\multicolumn{1}{c}{single operation}  &\multicolumn{1}{c}{multi-step} \\\midrule
    En & Keith has 20 books . Jason has 21 books . How many books do they have together ?& Lisa flew 256 miles at 32 miles per hour. How long did Lisa fly? & A chef needs to cook 9 potatoes. He has already cooked 7. If each potato takes 3 minutes to cook, how long will it take him to cook the rest?\\
    Zh & 基思有20本书。杰森有21本书。他们总共有多少本书？ & 丽莎以每小时32英里的速度飞行了256英里。丽莎飞了多长时间？ & 厨师需要煮9个土豆。他已经煮了7个了。如果每个土豆煮3分钟，剩下的他要煮多久？ \\\midrule
    Size & \multicolumn{1}{c}{395} & \multicolumn{1}{c}{562}& \multicolumn{1}{c}{600} \\
    \bottomrule
    \end{tabular}
    }
    \caption{Examples from each dataset used for zero-shot cross-lingual evaluation.
    % {\color{blue}shrink the table to make more writing space}
    }
    % \space
    \label{tab:examples}
\end{table*}

\subsection{Zero-shot Cross-lingual Evaluation Datasets} 
\label{subsec:bilingual}

To test cross-lingual transferability of MWP solvers, we make use of problem-type-specific datasets discussed in Section~\ref{subsec:datasets} as evaluation datasets, including  AddSub~\cite{hosseini-etal-2014-learning}, SingleOp~\cite{roy-etal-2015-reasoning} and MultiArith~\cite{roy-roth-2015-solving}.
To extend these datasets for cross-lingual evaluation, we use online machine translation APIs to translate them into Chinese and further manually refine the translations to be more native.
For each dataset, we list an example in Table~\ref{tab:examples}, in both English~(En) and Chinese~(Zh).

\section{Template-based Contrastive Training}
Math word problems can be categorized by expression templates if we replace numerical values of expressions with a special token.
Such templates have been adopted for supervision in other math solver approaches like \cite{wang-etal-2018-translating} and \cite{xie-etal-ijcai-2019-goal}.
Different from these methods, we don't use templates directly for supervision but make an assumption that problems sharing the same template are closer with each other from the point view of arithmetics, regardless of the surface forms of languages and descriptions.

To make use of this assumption, we introduce inter-language template-based contrastive training into our training process.
Specifically, we first group math word problems based on their templates.
During training, we pair each problem with a random sample from a different language sharing the same template.

As the representation learned by the encoder in Section~\ref{sec:method} is $\mathbf{M}$, we use its maxpooling with normalization as the latent representation for each problem and its positive sample, denoted as $\mathbf{z}$ and  $\mathbf{z}^+$ respectively.
Then, we conduct a batch-level contrastive training similar to SimCLR~\cite{pmlr-v119-chen20j} and use the NT-Xent loss (the normalized
temperature-scaled cross entropy loss) as the following:
\begin{equation}
% \footnotesize
\small
    \mathcal{L} = 
    -\log\frac{\exp(\langle\mathbf{z}, \mathbf{z}^+\rangle/\tau)}{\exp(\langle\mathbf{z}, \mathbf{z}^+\rangle/\tau\rangle + \sum_{j=1}^{N-1}\exp(\langle\mathbf{z}, \mathbf{z}_j^-\rangle/\tau)},
\end{equation}
where $\langle\cdot,\cdot\rangle$ is the inner product of the two vectors and the batch size is $N$.

It's worth noting that the distribution of templates is highly skewed.
In our experiments, we further consider two settings:
(1) \textbf{CL}, contrastive learning, when a problem doesn't have a candidate with the same template from another language, it contrasts with itself.
(2) \textbf{CL + TC}, contrastive learning with template constraint, we only use those problems which have at least one sample from another language.

Our contrastive learning approach differs with that of \citet{li-etal-2022-seeking} in the following ways: 
(1) our method is focusing on cross-lingual setting that each pair of examples come from different languages,
(2) we use batch-level contrastive training in consist with SimCLR.

There are also other works making use of latent representations of math word problems to enhance generalization ability of math solvers.
For example, \citet{liang2021solving} designed a teacher module to make
the latent vector to match the correct solution rather than its variations.
\section{Experiments}
\subsection{Experiment Setup}

\begin{table*}[ht]
    \centering
    \begin{tabular}{lrrrrrrrrrrrrr}\toprule
    \multirow{4}{*}{Model} & &\multicolumn{2}{c}{Test} & &\multicolumn{8}{c}{Zero-shot} \\\cmidrule{3-4}\cmidrule{6-13}
    & &Math23K &MathQA & &\multicolumn{2}{c}{AddSub} & &\multicolumn{2}{c}{SingleOp} & &\multicolumn{2}{c}{MultiArith} \\\cmidrule{3-4}\cmidrule{6-7}\cmidrule{9-10}\cmidrule{12-13}
    & &zh &en & &zh &en & &zh &en & &zh &en \\\midrule
    mBERT-zh & &\textbf{76.5} &3.3 & &30.9 &10.4 & &66.0 &32.7 & &\textbf{51.2} &15.7 \\
    mBERT-en & &0.5 &77.9 & &2.8 &6.1 & &5.0 &10.5 & &5.0 &3.2 \\
    \midrule
    XLM-R-xl & &75.5 &\textbf{79.0} & &\textbf{39.0} &21.3 & &67.4 &40.4 & &44.7 & \textbf{18.3} \\
    mBERT-xl & &76.3 &\textbf{79.0} & &35.2 &\textbf{24.1} & &\textbf{69.8} &\textbf{41.6} & &45.0 &16.0 \\
    \bottomrule
    \end{tabular}
    \caption{Comparisons of different cross-lingual models over Test set and zero-shot datasets. }
    \vspace{-3mm}
    \label{tab:xling}
\end{table*}

\paragraph{Evaluation metrics:} The model is expected to be a math problem solver, so the generated expressions should be executable by a specific compiler and executor.
During evaluation, each problem is counted as solved if the absolute error rate for the executed value and the target value is lower than a predefined threshold.
In our experiments, we choose $1e^{-4}$ as the threshold.
The final evaluation metric is the accuracy of solved problem against all the problems.

\paragraph{Methods to be compared:} We empirically compare the following cross-lingual: (1) \textbf{mBERT-zh} is using original multilingual BERT~\cite{devlin-etal-2019-bert} but trained over Math23K only; (2) \textbf{mBERT-en} is using original multilingual BERT~\cite{devlin-etal-2019-bert} but trained over the adapted MathQA only, and multilingual methods: (1) \textbf{mBERT-xl} is using original multilingual BERT~\cite{devlin-etal-2019-bert} but trained by mixing Math23K and the adapted MathQA; (2) \textbf{XLM-R-xl} is using XLM-R~\cite{conneau-etal-2020-unsupervised} but trained by mixing Math23K and the adapted MathQA.

\paragraph{Other experiment settings:} 
We choose to use  multi-lingual BERT~(mBERT)~\cite{devlin-etal-2019-bert} for cross-lingual training.
We train our models using one Nvidia 2080ti and a batch size of 160. 
The learning rate is set to $3e^{-5}$ with a scheduler supporting polynomial decay.
The training lasts for at most 150 epochs and will stop after 30 epochs if no improvement is observed.\footnote{\url{https://github.com/VisualJoyce/AnDuShu}}

\subsection{Results}

We list experiment results of all the methods in Table~\ref{tab:xling}.

\paragraph{Cross-lingual MWP Solver}
The first research question we want to answer is to what extent a MWP solver trained on one language can work on another language, with the help of pre-trained multilingual language models.
Table~\ref{tab:xling} shows that the MWP solvers trained using either Math23K~( \textbf{mBERT-zh}) or MathQA~(\textbf{mBERT-en}) have achieved impressive performance when tested in the same language.
However, the performance over a different language drops drastically and is almost negligible.
In a word, the MWP solver is almost non-transferable when it is trained on one language but evaluated over a second with the same operator set.

\paragraph{Multilingual MWP Solver}
The second research question we want to answer is whether training a MWP solver on multiple languages helps improve its effectiveness compared with training on a single language.
We can see that mixing two languages to train can give us a more language-agnostic model as the performance on Test split of both languages are competitive with monolingual cases.
What's more, on the newly extended bilingual datasets, there are consistent improvements for most of the datasets, especially for the English language.

Considering the difficulty of problems, these bilingual evaluation datasets are closer to Math23K~(primary school) than to MathQA~(GRE or GMAT).
Adding that mBERT-zh is also doing better than mBERT-en on English language, we suspect domain similarity is more important than language for MWP solvers.
% \leicomment{ Guess that for readers who are not familiars with MWP solving, it might not be clear why MathQA is not close to the extended zero-shot datasets but Math23k close }
% which can better generalize to other domains.
% \lx{it's really about domains? or just languages?}
% }

\begin{table*}[ht]
    \centering
    % \begin{adjustbox}{max width=\textwidth}
    % \begin{small}
    % \scalebox{0.9}{
% \begin{tabular}{lrrrrrrrrrr}\toprule
% &\multicolumn{4}{c}{zh} & &\multicolumn{4}{c}{en} \\\cmidrule{2-5}\cmidrule{7-10}
% &Test &AddSub &SingleOp &MultiArith & &Test &AddSub &SingleOp &MultiArith \\\midrule
% mBERT-zh &\textbf{75.0} &32.9 &63.0 &\textbf{50.5} & &3.0 &13.7 &31.3 &17.8 \\
% mBERT-en &1.7 &1.8 &9.1 &2.0 & &77.1 &5.1 &18.1 &2.3 \\
% mBERT-xl &74.5 &\textbf{38.7} &63.0 &45.8 & &\textbf{78.4} &\textbf{20.8} &\textbf{31.7} &\textbf{20.0} \\
%     \bottomrule
\begin{tabular}{lrrrrrrrrrrrrr}\toprule
\multirow{4}{*}{Model} & &\multicolumn{2}{c}{Test} & &\multicolumn{8}{c}{Zero-shot} \\\cmidrule{3-4}\cmidrule{6-13}
& &Math23K &MathQA & &\multicolumn{2}{c}{AddSub} & &\multicolumn{2}{c}{SingleOp} & &\multicolumn{2}{c}{MultiArith} \\\cmidrule{3-4}\cmidrule{6-7}\cmidrule{9-10}\cmidrule{12-13}
& &zh &en & &zh &en & &zh &en & &zh &en \\\midrule
% XLM-R-xl & &75.5 &79.0 & &39.0 &21.3 & &67.4 &40.4 & &44.7 & 18.3 \\
mBERT-xl & &76.3 &79.0 & &35.2 &\textbf{24.1} & &\textbf{69.8} &41.6 & &45.0 &16.0 \\
\midrule
% mBERT-xl consistency & &78.1 &78.5 & &46.3 &19.0 & &68.3 &32.2 & &38.2 &14.3 \\
% mBERT-xl consistency & &77.5 &78.4 & &43.5 &17.7 & &63.9 &21.9 & &47.8 &11.8 \\
mBERT-xl + CL & &\textbf{79.4} &\textbf{79.4} & &39.5 &19.2 & &62.6 &29.9 & &\textbf{46.2} &10.8 \\
mBERT-xl + CL + TC & &72.4 &49.5 & &\textbf{44.8} &19.2 & &67.4 &\textbf{44.5} & &45.5 &\textbf{17.3} \\
\bottomrule
\end{tabular}
    % }
    % \end{small}
    % \end{adjustbox}
    \caption{Performance of template-based contrastive training models over Test set and zero-shot datasets. }
    \vspace{-3mm}
    \label{tab:contrast}
\end{table*}

\paragraph{Template-based Contrastive Training}

The last section of Table~\ref{tab:contrast} shows how contrastive learning affects performance.
Firstly, adding contrastive learning can further boost performance on the test set of both languages. 
There's a significant increase~(3 points) for Math23K.
However, in zero-shot evaluation settings, performance over English drops consistently.
We suspect this might be caused by the diversity of templates on MathQA is much larger than that of Math23K.

Therefore, we further conduct a template-constrained experiment that ensures each template can be found on both languages.
Due to the number of training cases are reduced, performance of the test sets also drop by a large margin.
However, English problems over zero-shot setting benefit most from this experiment, which further verifies that math word problems depend closely on the problem types of the training set.

\section{Conclusion}

In this paper, we revisit the math word problems using a generation-based method constructed over pretrained multilingual models.
To assist analysis of cross-lingual properties of math solvers, we adopt two large-scale monolingual datasets and further adapts MathQA into the same annotation framework with Math23K.
We also reuse earlier smaller datasets and upgrade them into bilingual datasets by machine translation and manual checking.
% Our experiments show that the transferability of math word problems subjects to many constraints like language and domain.
Our experiments show that the MWP solvers may not be transferred to a different language even if the target expressions have the same operator set and constants.
But for both cross-lingual and multilingual cases, it can be better generalized if problem types exist on both source language and target language.
% \lx{may be more interesting to give some concrete results here.}
Problems considered to be easy by humans may still be hard for a math solver trained over the same language but from a different domain.
This tells us that for math word problem solvers, it might be beneficial to consider balancing different question types and permitted operators during training.

\bibliography{anthology,custom}
\bibliographystyle{acl_natbib}

\appendix

\section{Method}
\label{sec:method}

\begin{figure*}[t]
\centering
\includegraphics[width=\linewidth]{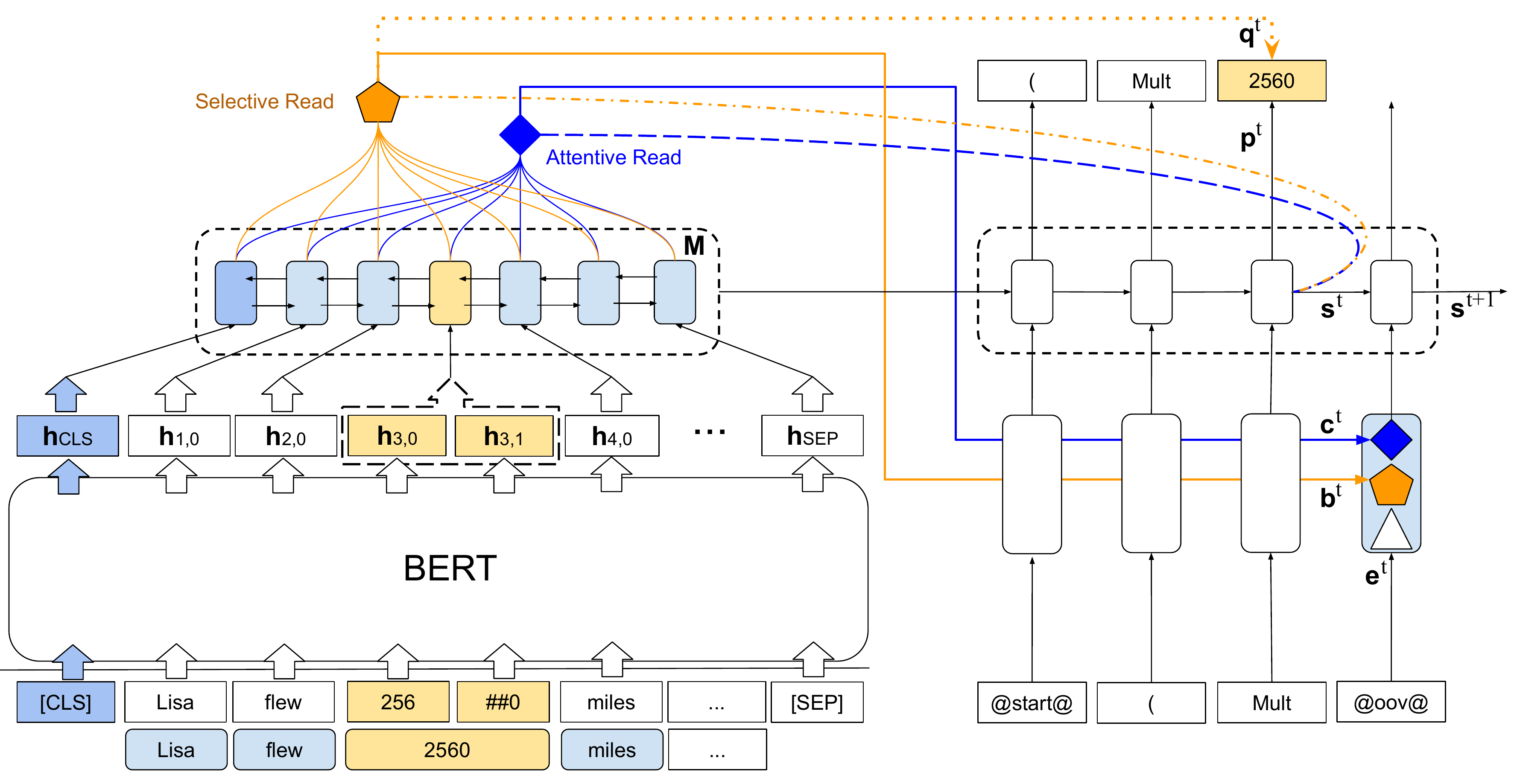}
\caption{Sequence-to-sequence model with copy mechanism.} 
\label{fig:model}
\end{figure*}

In this section, we construct a generation-based MWP solver using a sequence-to-sequence model with copy mechanism.
Our whole model can be visualized in modules through Figure~\ref{fig:model}.
The detailed illustration for each module is given as following:

\paragraph{Encoder} Our encoder is built upon a pretrained multilingual transformer, either BERT or XLM-R.
Suppose our input word $w_i$ is tokenized into word pieces  $(x_{i1}, x_{i2}, \ldots)$ and let $\mathbf{h}_{ij} \in \mathbb{R}^{d_h}$ denotes the hidden vector produced by the pretrained model representing $x_{ij}$.
We use average pooling to get the representation for the word $w_i$, denoted as $\mathbf{h}_i$.
Then we feed this contextualized representation of the math word problem into a two-layer bidirectional LSTM.
The output of this biLSTM is the encoder hidden states for decoding, denoted as $\mathbf{M} = (\mathbf{m}_0, \mathbf{m}_1, \ldots, \mathbf{m}_n) $.

\paragraph{Decoder} We use a LSTM cell as the decoding cell to predict the next token.
For each decoding step $t$, the cell will accept the embedding for previous word as input and output a decoder state $\mathbf{s}^t\in\mathbb{R}^{d_s}$.
Most of the numerical values in MWPs do not exist in the target vocabulary.
Therefore, we need copy mechanism~\cite{gu-etal-2016-incorporating} to facilitate generation of numerical values during decoding.
The copy scores are calculated as follows,
\begin{equation}
    u_i^t = \sigma(\mathbf{m}_i^\intercal\mathbf{W}_c)\mathbf{s}^t
\end{equation}
where $\mathbf{W}_c\in\mathbb{R}^{d_h\times d_s}$.
However, the embedding of a copied token will be identical to an out-of-vocabulary token.
To better capture the information from last decoding step, we use the copy score to further derive a state of selecting from source tokens, which is called \emph{Selective Read}.
\begin{align}
    \mathbf{q}^t &= \mathrm{softmax}(\mathbf{u}^t) \\
    \mathbf{b}^t &= \sum_{i} q^t_i\mathbf{m}_i
\end{align}

We use a bilinear attention to attentively read information from $\mathbf{M}$, getting the context vector $\mathbf{c}^t$, which is called \emph{Attentive Read}.
\begin{align}
    v_i^t &= \sigma(\mathbf{m}_i^\intercal\mathbf{W}_a\mathbf{s}^t + b) \\
    \mathbf{d}^t &= \mathrm{softmax}(\mathbf{v}^t) \\
    \mathbf{c}^t &= \sum_{i} d_i^t\mathbf{m}_i
\end{align}
where $\mathbf{W}_a\in\mathbb{R}^{d_h\times d_s}$.

From the problem definition, the target vocabulary is $\mathcal{V}=\mathcal{O}\cup\mathcal{C}$.
The generation score for the next token is given by:
\begin{equation}
    \mathbf{p}^t = \mathbf{W}_d^\intercal\mathbf{s}^t + b
\end{equation}
where $\mathbf{W}_d\in\mathbb{R}^{d_s\times |\mathcal{V}|}$.

The state updating process for the decoding cell takes in a fused information of last word embedding $\mathbf{e}^t\in\mathbb{R}^{d_e}$, selective read state $\mathbf{b}^t$ and attentive read state $\mathbf{c}^t$.
\begin{equation}
    \mathbf{s}^{t+1} = f(\mathbf{W}_s[\mathbf{e}^t,\mathbf{b}^t,\mathbf{c}^t], \mathbf{s}^{t})
\end{equation}
where $\mathbf{W}_s\in\mathbb{R}^{d_s\times (d_e+d_h+d_h)}$.

\begin{table*}[t]
    \centering
    % \begin{adjustbox}{max width=\textwidth}
    % \begin{small}
    \begin{tabular}{p{0.5\linewidth}p{0.5\linewidth}}\toprule
    Adapted Operators & Filtered Operators \\\midrule
    \begin{verbatim}
add,subtract,multiply,
rectangle_area,divide,
speed,power,negate,inverse,
square_area,sqrt,
square_edge_by_area,
cube_edge_by_volume,
volume_cube,surface_cube,
square_perimeter,
rectangle_perimeter,
stream_speed,triangle_area,
triangle_perimeter,surface_sphere,
volume_sphere,rhombus_area,
quadrilateral_area,volume_cylinder,
circle_area,volume_cone,circumface,
diagonal,volume_rectangular_prism,
original_price_before_loss,
original_price_before_gain,
p_after_gain,
square_edge_by_perimeter,negate_prob
    \end{verbatim} 
    & 
    \begin{verbatim}
floor,choose,min,tangent,sine,
reminder,lcm,factorial,gcd,max,
permutation,
triangle_area_three_edges,
surface_cylinder,rhombus_perimeter,
surface_rectangular_prism,
speed_in_still_water,log 
    \end{verbatim}\\
    \bottomrule
    \end{tabular}
    % \end{small}
    % \end{adjustbox}
    \caption{Operators that are adapted in MathQA.}
    \label{tab:operators}
\end{table*}

% This is an appendix.
\end{CJK*}

\end{document}